\pdfoutput=1

\documentclass[11pt]{article}
\usepackage{ACL2023}
\usepackage{amsfonts}
\usepackage{times}
\usepackage{latexsym}

\usepackage[T1]{fontenc}

\usepackage[utf8]{inputenc}

\usepackage{microtype}

\usepackage{inconsolata}

\usepackage[draft,textsize=footnotesize,textwidth=15mm]{todonotes}
\usepackage{multirow}

\usepackage{xcolor}
\usepackage{amsmath}
\usepackage{caption}
\usepackage{subcaption}

\newcommand{\m}{{\textbf{J-Guard}}}

\title{J-Guard: Journalism Guided Adversarially Robust Detection of AI-generated News}

\author{Tharindu Kumarage ~ Amrita Bhattacharjee ~ Djordje Padejski ~  Kristy Roschke \\ \textbf{Dan Gillmor} ~  \textbf{Scott Ruston} ~ \textbf{Huan Liu} ~ \textbf{Joshua Garland} \\
         Arizona State University \\
        \texttt{\{kskumara,abhatt43,padejski,carver,dan,scott.ruston,huanliu,jtgarlan\}@asu.edu}}
        
\begin{document}
\maketitle
\begin{abstract}

The rapid proliferation of AI-generated text online is profoundly reshaping the information landscape. Among various types of AI-generated text, AI-generated news presents a significant threat as it can be a prominent source of misinformation online. While several recent efforts have focused on detecting AI-generated text in general, these methods require enhanced reliability, given concerns about their vulnerability to simple adversarial attacks. Furthermore, due to the eccentricities of news writing, applying these detection methods for AI-generated news can produce false positives, potentially damaging the reputation of news organizations. To address these challenges, we leverage the expertise of an interdisciplinary team to develop a framework, {\m}, capable of steering existing supervised AI text detectors for detecting AI-generated news while boosting adversarial robustness. By incorporating stylistic cues inspired by the unique journalistic attributes, {\m} effectively distinguishes between real-world journalism and AI-generated news articles. Our experiments on news articles generated by a vast array of AI models, including ChatGPT (GPT3.5), demonstrate the effectiveness of {\m} in enhancing detection capabilities while maintaining an average performance decrease of as low as 7\% when faced with adversarial attacks.

\end{abstract}

\section{Introduction}

Recent advances in transformer-based generative models have led to substantial enhancements in the Natural Language Generation (NLG) capabilities of advanced conversational AIs, such as ChatGPT and BARD. These AI tools generate human-like text on a large scale by leveraging state-of-the-art (SOTA) pre-trained language models (PLMs) such as GPT 4 \cite{openai2023gpt4}, GPT 3.5 \cite{ouyang2022training}, GPT 3 \cite{radford2019language}, OPT \cite{zhang2022opt} and Lambda \cite{thoppilan2022lamda}. Considering the current trend of deploying these models in services offered to the general public, we can anticipate further improvements in NLG from future models.

However, deploying such NLG-capable models for public use poses the risk of potential misuse. Adversaries can employ these models to establish harmful agendas and conduct influence operations that deceptively steer the opinions of large groups of a target populace \cite{shu2020combating, goldstein2023generative}. AI-generated news articles are particularly concerning, as they can cause significant damage to the information ecosystem. Malicious actors can easily prompt AI models to generate text that purports to be authentic news but contains falsified information \cite{shu2018deep, zellers2019defending}. To make matters worse, current models are capable of generating misinformation and factually incorrect text in large volumes at a minimal cost through APIs. A recent report \footnote{https://www.newsguardtech.com/special-reports/newsbots-ai-generated-news-websites-proliferating/} by NewsGuard, an organization that combats misinformation online, identified an emerging set of 49 newsbots, i.e., news and information sites, that appear to incorporate AI for news generation. Therefore, it is crucial to have computational methods to discern between AI-generated news and actual human-written news to combat the persistent challenges to the information ecosystem.



%

In recent years, much interesting work has been done on detecting AI-generated text \cite{zellers2019defending, mitchell2023detectgpt, kirchenbauer2023watermark}. However, most of these methods, which we discuss in our Related Works section, do not explicitly focus on AI-generated news. Therefore, using these general-purpose AI text detectors to detect AI-generated news has a few challenges: 1) the unique attributes of professional journalism make news articles distinct from typical human-written text. Thus applying general AI text detection methods for AI-generated news detection could lead to false positives that potentially damage the reputation of journalists and news organizations, and 2) existing AI text detectors are highly vulnerable to adversarial attacks, e.g., paraphrasing\cite{sadasivan2023can,krishna2023paraphrasing}. 


%
To address the above challenges, we leverage the expertise of an interdisciplinary team, which includes journalists, computer scientists, and communication scholars, to develop a framework for \textbf{J}ournalism \textbf{Gu}ided \textbf{A}dversarially \textbf{R}obust \textbf{D}etection of AI-generated News ({\m}).
%
To this end, we first studied the unique professional journalism attributes of human-written news articles' writing and publishing process. Throughout the journalism process, many stylometric cues are incorporated, including journalism standards employed by the journalist as well as specific newsroom style guides and standards imposed by the newsroom editors. Here we hypothesize that even though the PLMs learn human-level writing via pretraining, they potentially will display semantic gaps in replicating these style guides and journalism standards inherent to the news production process. Therefore, we propose incorporating a simple yet effective set of auxiliary stylistic cues to guide the existing supervised AI text detectors to discern real-world journalism with the AI generation of news articles using PLMs. Furthermore, as we will show, since these cues quantify the high-level stylometry of the text, the detection process is more robust to the character and word level perturbations, thus, reinforcing the adversarial robustness of our AI-generated news detection methodology. 

To summarize, the main contributions of our paper are as follows: 
 \begin{enumerate}
	 \item To the best of our knowledge, we are the first to study and quantify stylistic cues resulting from the latent journalism process in real-world news organizations towards discriminating AI-generated news. 
	 \item We propose a computational framework incorporating these stylistic cues to detect AI-generated news.
	\item We conduct extensive experiments on a publicly available vast array of PLMs, including ChatGPT (GPT 3.5), to show our approach's effectiveness in detecting AI-generated news.
	\item By producing character and word level attacks, we empirically show how the stylistic cues we incorporated improved the adversarial robustness of AI-generated news detection.
\end{enumerate}

\section{Journalism Background}
\label{sec:jouranalysis}

Journalism as an industry does not universally subscribe to codes of conduct, owing largely to a historical rebuke of standardization as a profession \cite{shapiro2010evaluating}. Several trade groups, including the Society for Professional Journalists, have created detailed style guides. Many news organizations have adopted them internally, and others have created their versions. Scholars \cite{broersma1880form, shapiro2010evaluating, mateus2018journalism} have noted that, though the reporting process is typically situational, which makes it difficult to routinize, there are some key areas in which common methods, processes, and values signal an intent to establish credibility. And the form and style are integral to convincing people of the ‘truthiness’ of newsworthy events \cite{broersma1880form, mateus2018journalism}.

Journalistic practices that have been widely adopted include the use of the inverted pyramid as a storytelling format \cite{mateus2018journalism} and a style of writing based on the Associated Press Stylebook \footnote{https://www.apstylebook.com/}. Mateus \cite{mateus2018journalism} describes form and style “as key components of journalistic discourse that, in a given time, are able to generate credibility and confidence.” Though the AP Stylebook is not universally followed among news organizations, and some make situational exceptions, if we encounter purported news articles that are widely divergent from what AP recommends, we hypothesize that this is a strong signal of inauthenticity. In fact, adherence to the Stylebook is one of the key factors in the Associated Press’ automated journalism efforts \cite{linden2017algorithms}.

\begin{figure}
     \centering
    \includegraphics[width=0.46\textwidth]{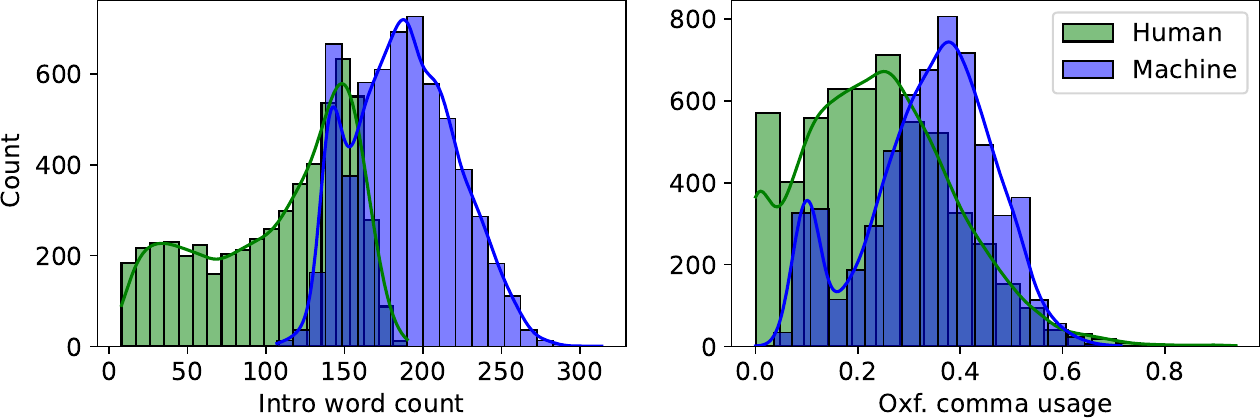}
     \caption{Distribution of GPT3 Generated News vs Human-written News.}
\label{fig:feat_dist}
\end{figure}

In our study, we aim to integrate the aforementioned hypothesis of inauthenticity into the task of detecting AI-generated news. Specifically, we investigate the extent to which current AI models are capable of generating news articles that adhere to professional journalism standards. Figure \ref{fig:feat_dist} illustrates a clear distinction in the distribution between GPT3-generated news articles and those written by humans from reputable news organizations such as CNN and the Washington Post. As illustrative examples of journalism features, we consider the length of introductory sections (leading sentences and paragraphs) and the usage of Oxford commas. Professional journalism typically employs shorter and more concise introductions, while the use of Oxford commas is infrequent in accordance with AP standards. Hence, we observe the potential for leveraging our hypothesis to enhance the detection of AI-generated news. In the subsequent section, we will delve into a detailed discussion of the journalism features that can be utilized for detecting AI-generated news.

\section{AI-generated News Detection}
This section presents the details of the {\m} framework. The {\m} framework consists of two main components: (a) the base AI text detector component and (b) the Journalism guidance component. The base AI text detector is any PLM sequence classification model. The journalism guidance component injects auxiliary journalism cues into the detection pipeline, thus transforming the base detector into an AI-generated news detector. We will provide a comprehensive discussion of these two components in the following sections.

\begin{figure}[h]
     \centering
     \includegraphics[width=0.4\textwidth]{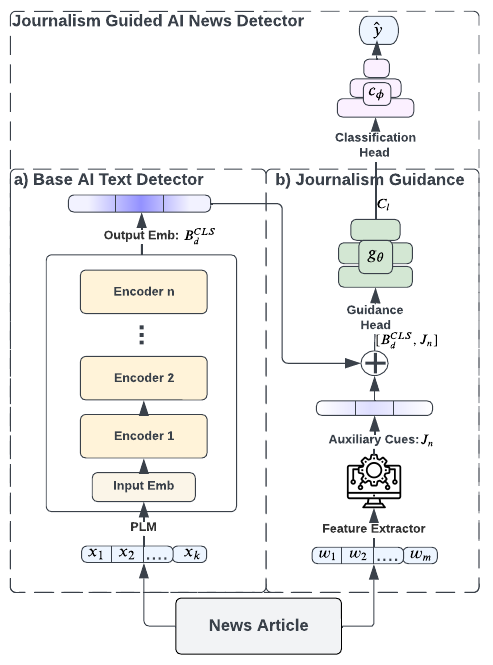}
     \caption{Proposed framework {\m}: The base detector component here is a supervised PLM-based detector for AI text detection.}
\label{fig:jguided_det}
\end{figure}

\subsection{Base AI Text Detector}

The Base AI text detector component consists of a pretrained transformer encoder stack with $n$ encoders to learn the semantic representation of the given input news article $X$. 
%
Here we define $(x_{1}, x_{2}, ..., x_{k})$ as the token representation of the input $X$ according to the tokenizer of the PLM model we choose for the base detector component. We denote the representation learned by the base AI text detector as $B_{k\times d}$ where $k$ is the sequence length (i.e., number of tokens of the input news article), and $d$ is the hidden state size of an encoder block. From the representation matrix, $B_{k\times d}$, we select the final hidden vector representation of the special token [CLS], $B_{d}^{CLS}$ as the feature vector for our task of detecting AI-generated news. Then $B_{d}^{CLS}$ is passed to the journalism guidance component for further processing. 

\subsection{Journalism Guidance}
\label{j_guidance}

The cornerstone of the {\m} framework lies in the journalism guidance of the Base AI text detector toward detecting AI-generated news—different modules within the journalism guidance component help achieve this goal. As illustrated in figure \ref{fig:jguided_det}, the journalism guidance component comprises a Journalism Feature Extractor and a Guidance Head as sub-components.

\subsubsection{Journalism Feature Extractor}
As postulated in Section~\ref{sec:jouranalysis}, encountering a news article that widely deviates from the recommended styles and standards of the AP stylebook may serve as a strong indication of inauthenticity. Therefore, the journalism feature extractor is a computational module that incorporates this hypothesis to enhance the detection of AI-generated news. The feature extractor takes the input news article $X$ in the form of a set of tokens $w_{1}, w_{2}, ..., w_{m}$. Here, $(w_{1}, w_{2}, ..., w_{m})$ represents the tokenized version of the input article $X$ using an improved Treebank Word Tokenizer\footnote{https://www.nltk.org/api/nltk.tokenize.word\_tokenize.html}. Subsequently, a set of extractor functions $f \in F$
is applied to these tokens to extract various scores that quantify the divergence of the input article from the AP recommended styles and standards. For discussion, it is useful to label three subsets of the extractor function set $F$  {$F^i| i \in {1, 2, 3}$}, such that $F=F^1 \cup F^2 \cup F^3$. The three subsets can be broadly defined as follows:

\begin{enumerate}
  \item $F^1$: \textbf{Organization and grammar standards} - functions that quantify the wording and grammatical structure of the news article (sentences and paragraphs forming) 
  \item $F^2$: \textbf{Punctuation usage} - functions that quantify the punctuation usage of the news article
  \item $F^3$: \textbf{Formatting standard violations} - functions that quantify the violations of the formatting of different elements in a news article, such as date, time, and number, in reference to the AP standards
\end{enumerate}
 
Within each extractor category, we extract multiple features that can quantify the deviation of the input article from the AP recommended styles and standards. In $F^1$, we examine the overall wording structure of the news article, as well as the leading sentence and paragraph, as the size of these components could serve as indicators of inauthenticity. For example, a large leading or introductory part is not very common for news articles. Additionally, we consider grammatical elements, such as tense and voice, which can provide cues about journalism standards. For instance, the use of past tense and passive voice is not common in news writing. As a result, the following features are extracted: mean word count (WC), mean sentence count (SC), WC of the leading sentence, WC of the leading paragraph, mean SC with passive voice, and mean SC with past tense.
In $F^2$, we analyze the usage of punctuation. In addition to standard punctuation marks, we also examine symbols that are rarely found in genuine news articles, such as the number sign. Consequently, the following features are extracted in $F^2$: mean usage of "$^^21$" "\#", "\textquotesingle", and the Oxford comma per paragraph.
Lastly, under $F^3$, we investigate format violations in the input article based on AP standards. Specifically, we identify and count violations related to date, time, and number formats. The detailed implementations of each feature extractor function can be found in the appendix section \ref{sec:appendix_fextrc}.

Some of the feature extractors mentioned above return mean values in the range of [0,1] while some return absolute counts, which will be larger than 1. Therefore we normalize the feature vector before incorporating it with the task of AI-generated news detection. Let $n$ be the number of journalism features, $f_i \in F$ and $W =( w_{1}, w_{2},\dots, w_{m})$ be the treebank tokenization of $X$ then we define the final normalized journalism feature vector $J_{n}$ as:

\begin{equation}
		J_{n} = \frac{[f_1(W), f_2(W), ..., f_n(W)]}{||[f_1(W), f_2(W), ..., f_n(W)]||}. 
\label{eq:feat_extrc}
\end{equation}

\subsubsection{Guidance Head}

We propose to enhance the detection capabilities and adversarial robustness of our detector by incorporating the learned journalism features, $J_{n}$, into the output of the base AI text detector, $B_{d}^{CLS}$. A naive approach would be to simply concatenate both features and pass them through the fully connected feedforward neural network, which we refer to as the Classification Head, to predict the final classification label $\hat{y}$. However, this naive approach may lead to the overshadowing of $J_{n}$ by $B_{d}^{CLS}$ due to the large dimensionality of $B_{d}^{CLS}$ compared to $J_n$. Furthermore, direct concatenation of the two feature vectors without considering their different ranges poses a feature scaling issue. To address these challenges, we propose the incorporation of an additional set of feedforward layers, referred to as the Guidance Head. This Guidance Head includes a hidden layer with a size equal to or larger than the input layer. This choice is made to prevent overshadowing of $J_{n}$. The Guidance Head learns the relationships between the feature vectors $B_{d}^{CLS}$ and $J_{n}$, without overshadowing $J_{n}$, by mapping the input [$B_{d}^{CLS}, J_{n}$] to a higher-dimensional feature space. Note that we first normalize the [$B_{d}^{CLS}$, $J_{n}$] vector before passing it to the Guidance Head to avoid feature scaling issues. Finally, the Guidance Head's output layer produces a reduced vector of the scaled-up hidden representation, which we pass to the Classification Head for the final prediction. To summarise, as shown in equation \ref{eq:fusion}, the whole purpose of Guidance Head is to learn the function $g_{\theta}$ that learns fusion between $B_{d}^{CLS}$ and $J_{n}$. 

\begin{equation}
	C_l = g_{\theta}\left(\frac{[B_{d}^{CLS}, J_{n}]}{||[B_{d}^{CLS}, J_{n}]||}\right)
\label{eq:fusion}
\end{equation}

Here, $C_l$ is the reduced vector of size $l$ produced by the output layer of the Guidance Head.


Finally, the output of the Guidance Head, $C_l$ is passed to the Classification Head to predict the final classification label $\hat{y}$. Using the ground truth label, we incorporate standard cross-entropy loss to train the whole framework {\m} end to end.

\section{Experiments and Results}

This section describes the experimental settings used to validate our framework, including the datasets and baselines, followed by a thorough analysis of the experiments. We conducted several experiments to investigate whether the proposed journalism features can improve the detection of AI-generated news. We aim to answer the following two research questions through our experiments:

\begin{itemize}
	\item \textbf{RQ1} - Do the identified journalism features, enhance the detection of AI-generated news?
	\item \textbf{RQ2} - Do the identified journalism features, enhance the adversarial robustness of AI-generated news detection?
\end{itemize}

\subsection{Datasets and AI Generators}
\label{datasets}


We evaluate our approach on a vast array of AI generators, i.e., PLMs --- To this end, we use the benchmark dataset TuringBench~\cite{uchendu2021turingbench}. TuringBench is a dataset consisting of human-written news articles, mostly from CNN and the Washington Post, and AI-generated news from more than 10 PLM generators. Of these, we used the following PLMs for our analysis:  Grover, CTRL, PPLM$_{gpt2}$ (base model used is GPT2), GPT2, GPT3. Within TuringBench, data is generated using various combinations of PLMs and model sizes. To maintain brevity in our analysis, we have included only the largest model size for each PLM. This selection is justified by the understanding that the largest model size for each PLM is expected to produce the highest quality text, making it more challenging to detect. Therefore, our results can be extrapolated to smaller PLMs as well.

Furthermore, we performed our experiments on a ChatGPT dataset that we created. Given the human-like quality of text generated by newer PLMs like GPT3.5 and GPT4~\cite{openai2023gpt4}, it is important to evaluate our detection framework on such language models. To create this dataset, we followed steps similar to the ones in the TuringBench paper~\cite{uchendu2021turingbench}. Specifically, we sampled around 9,000 news articles from CNN and the Washington Post and use these as `human' written articles. For each of these articles, we prompt ChatGPT (with backend gpt-3.5-turbo, model version as of March 14, 2023) to generate an equivalent news article. To do this, we experiment with several types of prompts, and for the final data generation, we use the prompt: ``Generate a news article with the <headline>.'', where <headline> is the headline from the corresponding human written article. For the ChatGPT generations, we set $top\_p$ to 1, $temperature$ to 0.5 and limit the length of the generated text to 1024 tokens. The final dataset contains 9k human-written and 9k ChatGPT-generated articles, which we divided into train, test, and validation splits (7:2:1) similar to TuringBench. We will release this dataset to the public upon acceptance of the paper (section \ref{chatgpt_data}). 


\subsection{Baselines}

Our experiments consist of two categories of AI news detector baselines: First, we study simple feature-based classification schemes which use logistic regression (LR) with BOW and Word2vec features as a baseline to evaluate the quality of the journalism features (JF) we selected via our journalism analysis. Second, we aim to empirically compare and validate {\m} with SOTA PLM-based methods for AI-generated text detection. The SOTA baselines can be further categorized into 1) \textbf{Zero-shot PLM-based classifiers}: GLTR~\cite{gehrmann2019gltr}, and the newer zero-shot baseline DetectGPT~\cite{mitchell2023detectgpt}. These two approaches work without supervised training datasets for detecting AI-generated text and 2) \textbf{Supervised PLM-based classifiers}: We consider OpenAI's GPT-2 detector (RoBERTa-large) as our supervised PLM-based detector baseline. We considered two variants of this model i) OpenAI$_{ Zero}$ - OpenAI's off-the-shelf GPT-2 detector without any task-specific tuning, ii) OpenAI$_{ FT}$ - OpenAI's GPT2 detector finetuned for AI news detection. Further technical details about the baselines can be found in the appendix section \ref{sec:appendix_base}.

\subsection{Detection Setup}
\label{det_setup}


\noindent \textbf{Implementation Details of {\m}:}
The base AI text detector is one of the key components of the {\m} framework, involving a supervised PLM specifically designed for detecting AI-generated text. In our research, we conducted experiments using various existing PLMs (base size), including RoBERTa, BERT, DeBERTa, and DistilBERT. Among these models, RoBERTa exhibited the highest performance, and therefore, it was selected as the base AI text detector for the {\m} framework, while reporting the experiment results. Both the Classification Head and the Guidance Head were implemented using feedforward neural networks comprising one hidden layer. For the training of the overall framework, a max length of 512, a learning rate of $2\times10^{-5}$, and a dropout rate of 0.2 were employed. The training process utilized a 40 GB NVIDIA A100 GPU ($\approx$ 1hr per AI generator).

\noindent \textbf{Task Details:}
We consider the task of AI-generated news detection as a binary classification problem. In our data, we have train, test, and validation (7:2:1 ratio) splits for each AI generator, where we use the train set to finetune models on the task of AI news detection and the test set to record the classification performance. The validation set was used for early stopping to determine the number of training epochs. See appendix section \ref{sec:appendix_imp} for more details. 

\subsection{Adversarial Attack Setup}
\label{ad_attack_setup}

In order to validate the adversarial robustness of the detector, we conducted two common attacks that have been observed in previous work: Cyrillic injection and paraphrasing ~\cite{crothers2022adversarial, sadasivan2023can, liang2023mutation}. In the Cyrillic injection attack, we perturbed the input text by replacing English characters with similar-looking Cyrillic characters. Specifically, we selected three highly frequent English vowels, "a", "e", and "o," and replaced them with their Cyrillic counterparts. For the paraphrasing attacks, we employed a PLM-based approach that incorporates the T5 model to paraphrase a given input text \cite{sadasivan2023can}. 

\subsection{Results and Discussion}
\label{re_dis}

This section discusses the experimental results under AI-generated news detection, including additional experiments on feature importance and PLM choice for the {\m}. Furthermore, we empirically show the adversarial robustness of the {\m} by emulating multiple attack scenarios. 

\subsubsection{RQ1 - AI-generated News Detection Performance}

Here, we present an evaluation of the performance of AI news detection using a wide range of AI generators. Table \ref{tab:detresults} reports the AUROC scores for different detectors (rows) across different AI generators/PLMs (columns). Based on the results in Table \ref{tab:detresults}, we make the following observations regarding AI-generated news detection: \\
\noindent 1) \textbf{Effectiveness of journalism features} - when we look at the logistic regression results (1st 3 result rows of Table~\ref{tab:detresults}), we can see that journalism features outperform simple BOW and word2vec performance across all the AI generators. This suggests that the journalism feature space provides a reasonable boundary for discriminating between human-written news and AI-generated news. \\
\noindent 2) \textbf{Effectiveness of {\m}} - Our proposed method outperforms all the detection baselines in 4 out of 6 AI generators. However, for PPLM$_{gpt2}$ and GPT2 generators, we observe that the finetuned OpenAI detector (OpenAI$_{FT}$) outperforms {\m} by a small margin. The OpenAI detector has an advantage in detecting GPT2 and PPLM$_{gpt2}$ as it is exposed to GPT2 samples in the first stage of finetuning done by OpenAI. \\
\noindent 3) \textbf{Effectiveness of task-specific training} - We observe that off-the-shelf zero-shot methods (GLTR, DetectGPT, and OpenAI$_{Zero}$) perform poorly across many AI generators in detecting AI news. However, the performance improves significantly when we further finetune the OpenAI$_{Zero}$ on the AI news detection task (OpenAI$_{FT}$). This observation highlights the importance of task-specific supervision.

\begin{table*}[t]
\centering
\begin{tabular}{|l|lllll|l|}
\hline
Dataset $\rightarrow$   & \multicolumn{5}{c|}{TuringBench}                                                                                                                                                                            & In-House Data                      \\ \hline
Generator $\rightarrow$ & \multicolumn{1}{l|}{\multirow{2}{*}{Grover}} & \multicolumn{1}{l|}{\multirow{2}{*}{CTRL}} & \multicolumn{1}{l|}{\multirow{2}{*}{PPLM$_{gpt2}$}} & \multicolumn{1}{l|}{\multirow{2}{*}{GPT2}} & \multirow{2}{*}{GPT3} & \multirow{2}{*}{ChatGPT} \\ \cline{1-1}
Detector  $\downarrow$                & \multicolumn{1}{l|}{}                        & \multicolumn{1}{l|}{}                      & \multicolumn{1}{l|}{}                      & \multicolumn{1}{l|}{}                      &                       &                                    \\ \hline
LR+ BoW                   & \multicolumn{1}{l|}{0.816}                   & \multicolumn{1}{l|}{0.775}                 & \multicolumn{1}{l|}{0.792}                 & \multicolumn{1}{l|}{0.822}                 & 0.806                 &       0.810                       \\ \hline
LR+ W2V                   & \multicolumn{1}{l|}{0.854}                   & \multicolumn{1}{l|}{0.793}                 & \multicolumn{1}{l|}{0.804}                 & \multicolumn{1}{l|}{0.871}                 & 0.852                 &        0.847                      \\ \hline
LR + JF                & \multicolumn{1}{l|}{0.897}                   & \multicolumn{1}{l|}{0.831}                 & \multicolumn{1}{l|}{0.873}                 & \multicolumn{1}{l|}{0.931}                 & 0.912                 & 0.883                              \\ \hline \noalign{\hrule height 1.5pt}
GLTR                    & \multicolumn{1}{l|}{0.482}                   & \multicolumn{1}{l|}{0.784}                 & \multicolumn{1}{l|}{0.634}                 & \multicolumn{1}{l|}{0.542}                 & 0.454                 &             0.728                       \\ \hline
DetectGPT                 & \multicolumn{1}{l|}{0.549}                   & \multicolumn{1}{l|}{0.806}                 & \multicolumn{1}{l|}{0.492}                 & \multicolumn{1}{l|}{0.505}                 & 0.557                 &                0.766                    \\ \hline 
OpenAI$_{Zero}$            & \multicolumn{1}{l|}{0.746}                   & \multicolumn{1}{l|}{0.763}                 & \multicolumn{1}{l|}{0.918}                 & \multicolumn{1}{l|}{0.857}                 &      0.773            &                     0.756        \\ \hline \noalign{\hrule height 1.5pt}
OpenAI$_{FT}$                   & \multicolumn{1}{l|}{0.975*}                   & \multicolumn{1}{l|}{0.969*}                 & \multicolumn{1}{l|}{\textbf{0.966}}                 & \multicolumn{1}{l|}{\textbf{0.980}}                 & 0.951*                 & 0.925*                             \\ \hline
\m          & \multicolumn{1}{l|}{\textbf{0.986}}                   & \multicolumn{1}{l|}{\textbf{0.972}}                 & \multicolumn{1}{l|}{0.965*}                 & \multicolumn{1}{l|}{0.975*}                 & \textbf{0.968}                 & \textbf{0.934}                              \\ \hline
\end{tabular}
\caption{Proposed {\m} model performance (AUROC) values for AI-generated news detection. Bold shows the best AUROC within each column (Detector-PLM generator combination); asterisk (*) denotes the second-best AUROC.}
\label{tab:detresults}
\end{table*}

\begin{figure}
     \centering
     \includegraphics[width=0.35\textwidth]{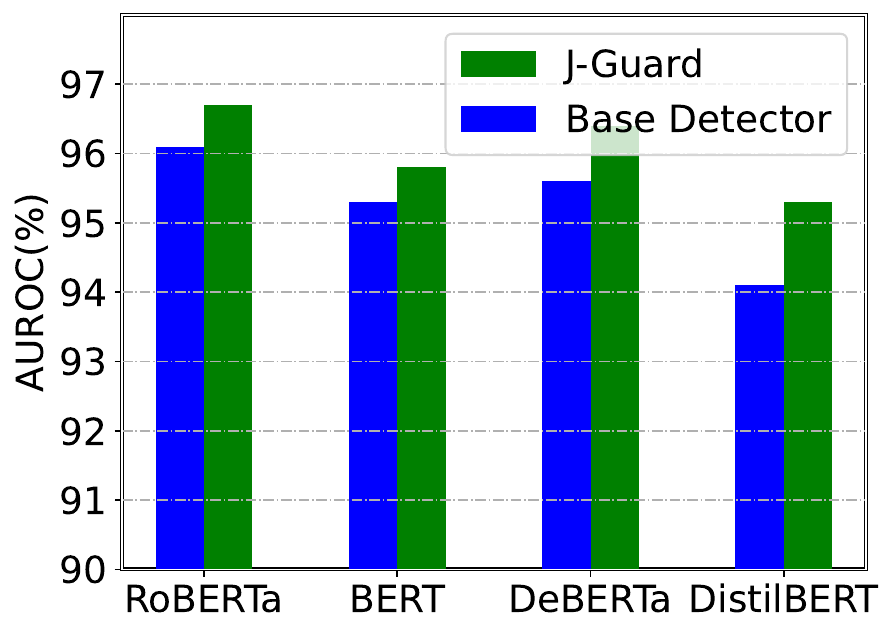}
     \caption{Effect of the choice of PLM for framework \m - Average AUROC across all six AI generators, before and after Journalism guidance.}
\label{fig:plm_choice}
\end{figure}

We also analyzed the impact of the base AI detector choice on our framework, {\m}. We experimented with multiple open-source PLMs, as shown in Figure \ref{fig:plm_choice}. We evaluate each PLM with and without {\m} to evaluate the detection performance and report the average performance across the AI generators considered in our study. We found that the detection performance could be enhanced with the use of {\m} on each PLM. Among all the models, RoBERTa yielded the best performance.

\begin{figure}
     \centering
     \includegraphics[width=0.45\textwidth]{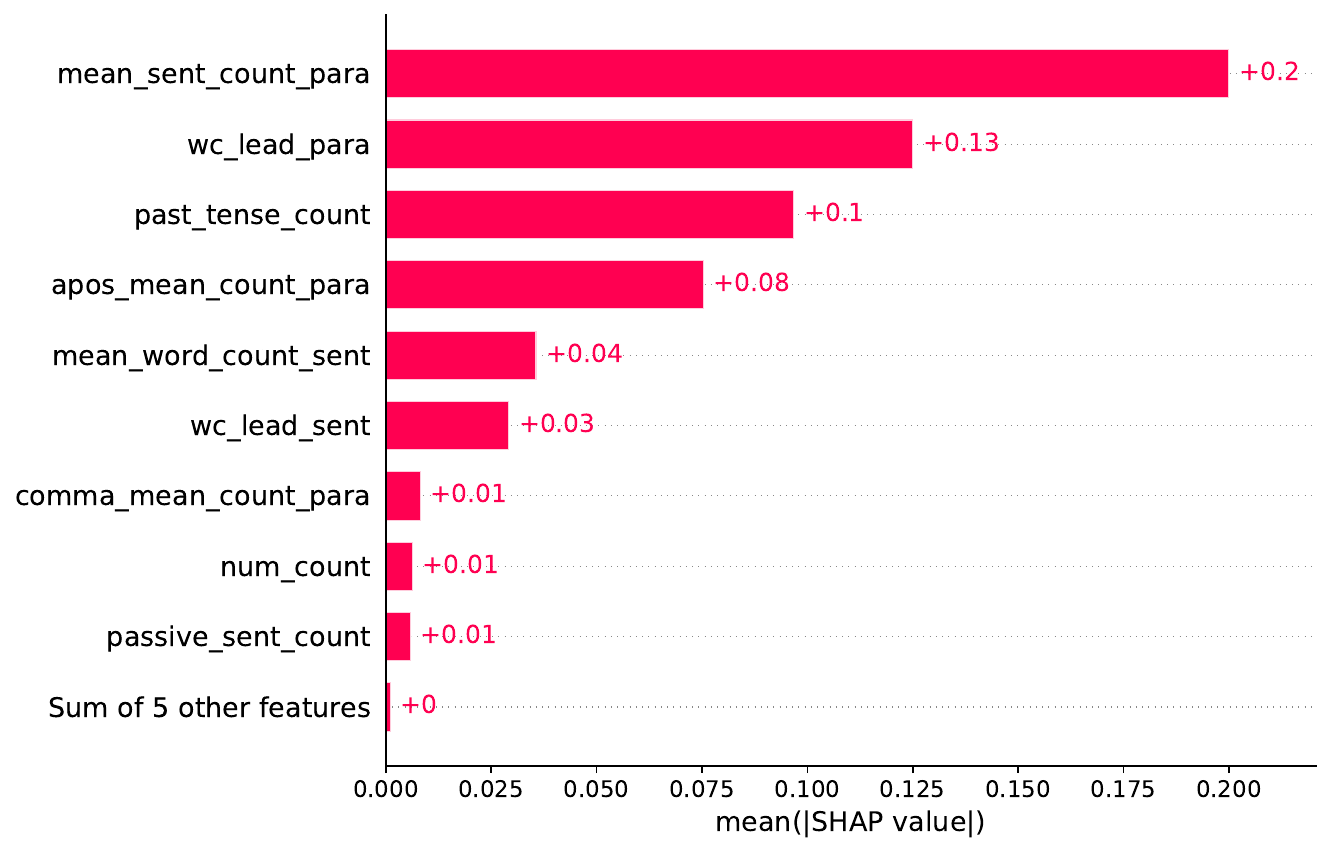}
     \caption{SHAP values to estimate journalism feature importance.}
\label{fig:feat_impo}
\end{figure}

Additionally, we conducted a study to better understand the significance of journalism features in AI news detection with the help of a Shapley Additive Explanations (SHAP) \cite{lundberg2017unified} explainer on the logistic regression classifier. The SHAP values were used to indicate feature importance. We only present SHAP plots for the GPT3 detection task for brevity reasons, but SHAP plots related to other AI generator detection tasks can be found in the appendix section \ref{sec:appendix_exp}. The SHAP plots show that certain features, such as mean sentence count for a paragraph (\textit{mean\_sent\_count\_para}), the word count of lead paragraph size (\textit{wc\_lead\_para}), and past tense usage (\textit{past\_tense\_count}) are highly significant in distinguishing AI news from human-written news, as depicted in Figure \ref{fig:feat_impo}.


\subsubsection{RQ2 - Adversarial Robustness of AI News Detection}
\label{rq2}

This section discusses our experiments on evaluating the adversarial robustness of AI news detectors. As outlined in section~\ref{ad_attack_setup}, we conducted two types of attacks on the detectors: character-level attacks involving Cyrillic injection and word-level attacks involving paraphrasing. Table \ref{tab:robust_results} shows the detectors' performance (AUROC) difference before and after each attack. For brevity, we only report the detection performance on GPT3 and ChatGPT data, while the other results can be found in the appendix section \ref{sec:appendix_exp}. Based on the results presented in Table \ref{tab:robust_results}, we make the following observations regarding the adversarial robustness of AI news detectors:

\noindent \textbf{Attack Success} - We have observed that almost every SOTA baseline detector we have considered is susceptible to adversarial attacks. On average, the performance of the detectors dropped by at least 15-20\%. In contrast, we observed a low attack success rate with the GLTR model. However, the observation of low attack success is meaningless as the GLTR model had a near-random guess ($\approx0.5$ AUROC) performance before the attack. 
The overall reduction in performance following the Cyrillic injection attack can be attributed to the tokenizer. Cyrillic letters in the input text alter the token representation, subsequently affecting detection. For paraphrasing, modifying the original text could alter the learned decision boundary during detector training, leading to a performance decline.

\noindent \textbf{Improved Adversarial Robustness of {\m}} - We have observed that {\m} is quite resilient to adversarial attacks, with an average performance drop of only 7\%. It is apparent that this robustness is due to the journalism features employed by \m. For example, OpenAI$_{FT}$, which shares the same PLM architecture and training data for detection as {\m}, has an average performance drop of nearly 15\%. 
In the journalism feature space, we check for high-level semantic gaps and violations of journalism standards. Character-level attacks, such as Cyrillic injection, have a negligible effect on these feature calculations. Even with paraphrasing attacks, the edit distance between the original and perturbed text may be substantial in the input space but insignificant in the journalism feature space, making {\m} robust to such attacks.

\begin{table}
\centering
\begin{tabular}{|l|ll|ll|}
\hline
Generator    $\rightarrow$          & \multicolumn{2}{c|}{GPT3}                                                     & \multicolumn{2}{c|}{ChatGPT}                                                  \\ \hline
Attack      $\rightarrow$     & \multicolumn{1}{l|}{\multirow{2}{*}{Para.}} & \multirow{2}{*}{Cyri.} & \multicolumn{1}{l|}{\multirow{2}{*}{Para.}} & \multirow{2}{*}{Cyri.} \\ \cline{1-1}
Detector $\downarrow$        & \multicolumn{1}{l|}{}                              &                          & \multicolumn{1}{l|}{}                              &                          \\ \hline
GLTR             & \multicolumn{1}{l|}{\textbf{0.041}}                         & 0.055                    & \multicolumn{1}{l|}{0.095}                         & 0.056                    \\ \hline
DetectGPT        & \multicolumn{1}{l|}{0.222}                         & 0.196                    & \multicolumn{1}{l|}{0.254}                         & 0.183                    \\ \hline
OpenAI$_{Zero}$  & \multicolumn{1}{l|}{0.244}                         & 0.201                    & \multicolumn{1}{l|}{0.223}                         & 0.154                    \\ \hline
OpenAI$_{FT}$   & \multicolumn{1}{l|}{0.159}                         & 0.138                    & \multicolumn{1}{l|}{0.166}                         & 0.150                     \\ \hline
{\m} & \multicolumn{1}{l|}{0.090}                         & \textbf{0.041}                    & \multicolumn{1}{l|}{\textbf{0.091}}                         & \textbf{0.040}                     \\ \hline
\end{tabular}
\caption{Detector performance change after the attack (AUROC before the attack - AUROC after the attack). Bold shows the lowest AUROC difference within each column (detector-attack combination).}
\label{tab:robust_results}
\end{table}

\section{Related Work}
\subsection{AI-generated Text Detection}




Several methods have been explored for detecting AI-generated text, such as logistic regression, SVC, etc.~\cite{ippolito2019automatic}. GLTR~\cite{gehrmann2019gltr} uses a set of simple statistical tests to check whether an input text sequence is AI-generated or not. Fine-tuned PLM detectors are also used and considered state of the art~\cite{solaiman2019release,jawahar2020automatic,zellers2019defending, kumarage2023stylometric}, such as OpenAI's GPT2 detector that uses a RoBERTa backbone finetuned with GPT-2 outputs ~\cite{radford2019language}. With the rapid advancement of newer language models like GPT3.5/4, there is a growing emphasis on the capabilities of few-shot or zero-shot detection and the interpretability of these detectors ~\cite{mitrovic2023chatgpt}. Some new detectors include commercial products such as GPTZero \footnote{https://gptzero.me/} and OpenAI's detector that is trained on the text generated by GPT-3\footnote{https://openai.com/blog/new-ai-classifier-for-indicating-ai-written-text}. An interesting zero-shot detection approach, DetectGPT~\cite{mitchell2023detectgpt}, operates on the hypothesis that minor rewrites of AI-generated text would exhibit lower log-probabilities under the model compared to the original sample. Watermarking ~\cite{kirchenbauer2023watermark} on PLM-generated text has gained attention as a detection mechanism in the research community. However, its success hinges on the cooperation and support of the organizations that develop the PLMs.

\subsection{Adversarial Robustness of AI Text Detection}

Multiple studies have examined the vulnerability of AI text detectors, specifically those designed for early PLMs like Grover and GPT2 \cite{crothers2022adversarial, liang2023mutation, gagiano2021robustness}. These studies conducted various attacks at the character and word levels, including flipping upper-lower case, using homoglyphs, misspelling words, and replacing synonyms. The results indicate that supervised-PLM-based AI text detectors are highly susceptible to these attacks, with success rates reaching up to 96\% in some cases \cite{crothers2022adversarial}. Recent research has also demonstrated that paraphrasing the input text can significantly undermine the performance of AI text detection approaches \cite{sadasivan2023can, krishna2023paraphrasing}, raising concerns about the reliability of such methods. Proposed solutions involve semantic retrieval to counter paraphrase attack \cite{krishna2023paraphrasing}, but they rely on text generation APIs like the OpenAI API, which limits their practical applicability when evaluating independent detection mechanisms.

Previous research has highlighted two important considerations for detecting AI-generated text. 1) it is impractical to rely on a single detector for all types of AI text, emphasizing the need for domain-specific models, and 2) ensuring the detector's robustness against adversarial attacks is critical, warranting further investigation in this field.

\section{Conclusion}

In this paper, we examine the task of detecting AI-generated news from a multidisciplinary perspective, aiming to identify domain-specific signals that can enhance detection accuracy while preserving robustness against adversarial attacks. We analyzed the real-world news production process compared to AI news generation and identified a set of stylistic cues that measure the deviation of AI-generated news from journalistic standards established by entities such as the Associated Press. Our proposed framework, {\m}, incorporated these auxiliary features and steered existing supervised PLM-based AI text detectors to achieve robust performance across various text-generation AIs, including ChatGPT. For future work, it would be interesting to see how prompt engineering can generate news articles that evade journalism-guided detection.  
 
\section{Limitations}

\subsection{Assumption of Professional Journalism}
In our study, we make the assumption that the human-written portion of the dataset is produced through a professional journalism process. This means that the news organization or journalist adheres to the journalism standards commonly defined by organizations such as the Associated Press (AP). It is important to note that our hypotheses and findings are valid only under this assumption. If the human-written articles come from a non-professional journalism source, we expect the detection performance to decrease since the distinction achieved through journalism features may no longer hold.

\subsection{Domain-Specific Training}
The approach we propose follows the supervised learning paradigm for AI news detection. As a result, it requires specific training data to be effective in real-world AI news detection scenarios. For instance, if we aim to ensure the performance of {\m} on an AI text generator $X$, we first need to gather a training dataset consisting of news articles generated by $X$. It is important to emphasize that our approach does not claim to have cross-AI generator generalized detection capabilities. However, the set of journalism features we proposed are agnostic to the AI generator and derived from real-world journalism process analysis.

\subsection{In-House Dataset}
As described in our section \ref{datasets}, we generated our dataset using ChatGPT due to the lack of publicly available ones. Although we followed a similar data collection and generation pipeline as TuringBench \cite{uchendu2021turingbench}, it is worth noting that there may be differences in the pre-processing and data cleanup we performed compared to the methods employed by the authors of TuringBench.

\subsection{Generalizability for ChatGPT-generated Text Detection}
Throughout our paper, we emphasize the specificity of our analysis and its focus on AI news detection. Therefore, the differences in ChatGPT text detection performance reported by the community \footnote{https://openai.com/blog/new-ai-classifier-for-indicating-ai-written-text}, as opposed to the high-performance results presented in our work, can be attributed to the domain of the data, specifically news articles. We hypothesize that detecting a particular domain, such as news articles with a specific type and text style, is easier than detecting generic text generated by ChatGPT. In summary, our paper does not claim that {\m} can be used for general ChatGPT text detection tasks; instead, it presents a specific method tailored to improve the detection of ChatGPT-generated news.

\section{Ethical Considerations}

\subsection{Intended Use}

It is crucial to consider the intended real-world application of {\m} and its societal impact. Our research on AI news detection aims to develop an algorithm that effectively identifies and mitigates the spread of misinformative, AI-generated news articles. The primary application of our work lies in online content moderation and forensics, where the decisions made by our detector can be utilized to flag or remove news articles from social media platforms, web search results, and other platforms. However, a significant ethical concern arises from potential false positives generated by our method. Suppose the detector incorrectly flags a genuine news article from a reputable organization as AI-generated. In that case, it may lead to the censorship of legitimate news, causing harm to the reputation and rights of the journalist and the publishing organization. Hence, we strongly advise users not to incorporate {\m} into fully automated real-world content moderation or forensics systems unless a human annotator or analyst works in conjunction with the system to make the final decision.

\subsection{ChatGPT-generated News}
\label{chatgpt_data}
In our study, we conducted experiments using the in-house ChatGPT-generated news articles. It is crucial to emphasize that we adhered to the usage policies\footnote{https://openai.com/policies/usage-policies} of OpenAI while generating these news articles through the API (refer to the prompt details in Section \ref{datasets}).
We recognize the importance of not publicly releasing any AI-generated news article, as we cannot guarantee the factual accuracy of the content. Therefore, we will implement an on-demand release structure for our ChatGPT-generated news articles. Individuals or organizations requesting access to our generated news articles for legitimate academic research purposes will be granted permission to download the data.

\subsection{Fairness and Bias in Detection}
Our research endeavors to prioritize using natural language processing tools for the betterment of society while upholding principles of fairness and impartiality. We transparently disclose our methodology, results, and, most importantly, limitations to mitigate biases and address ethical concerns. Furthermore, we commit to continuous assessment and improvement of our system in the future.

\subsection{Malicious Use of Adversarial Attacks}

We understand the potential danger of an adversary misusing the adversarial attack setup we presented in our section \ref{ad_attack_setup} to attack existing commercial AI text detectors. However, we posit that finding these limitations and vulnerabilities in AI text detector systems (red-teaming) will outweigh the potential for misuse, given we help future researchers mitigate these issues. 
However, as a precaution, we will not release the adversarial setup code base to the public. Similar to ChatGPT data, individuals or organizations requesting access to our adversarial attack setup for legitimate academic research purposes will be granted permission to receive the code base.

\bibliography{references}
\bibliographystyle{acl_natbib}

\appendix

\section{Feature Extractors}
\label{sec:appendix_fextrc}

In Section \ref{j_guidance}, we divided the feature extractor function set $F$ into three subsets: {$F^i| i \in {1, 2, 3}$}, where $F=F^1 \cup F^2 \cup F^3$. 

\begin{enumerate}
  \item $F^1$: \textbf{Organization and grammar standards} - functions that quantify the wording and grammatical structure of the news article, including sentences and paragraphs.
  \item $F^2$: \textbf{Punctuation usage} - functions that quantify the punctuation usage within the news article.
  \item $F^3$: \textbf{Formatting standard violations} - functions that quantify violations of formatting standards for various elements in a news article, such as date, time, and numbers, following AP standards.
\end{enumerate}

In this appendix section, we provide implementation details for the feature extractor functions in the categories mentioned above to facilitate the reproducibility of our framework, {\m}.

\subsubsection{$F^1$: \textbf{Organization and Grammar Standards}}

We used two NLTK tokenizers for our implementations: word\_tokenize for word-level tokenization and sent\_tokenize for sentence-level tokenization. 

\textbf{Mean Word Count (WC)}: We calculated the average word count per sentence in the input news article. First, we extracted the sentences using sent\_tokenize, and then for each sentence, we obtained the word tokens. The word count for each sentence was determined by counting all word tokens that contained at least one alphabetical character.

\textbf{Mean Sentence Count (SC)}: We counted the number of sentences per paragraph in the news article. To obtain the paragraphs, we split the input article using the newline character. For each paragraph, we obtained sentence-level tokens. The sentence count for each paragraph was determined by counting all sentences that contained at least one alphabetical character.

\textbf{Word Count of the Leading Sentence}: Here we focused on the first sentence, which is the lead sentence of the input news article. After obtaining the sentence-level tokens using sent\_tokenize, we calculated the word count for this leading sentence using the same approach as the Mean Word Count (WC).

\textbf{Word Count of the Leading Paragraph}: Similar to the previous approach, we extracted all paragraphs by splitting the input article based on the newline character. We then counted the number of word tokens in the first paragraph.

\textbf{Mean Sentence Count with Passive Voice}: Here we determined the number of sentences in a paragraph written in the passive voice. To identify the voice of a sentence, we employed a simple test. First, we extracted the dependency tree relations using the spaCy dependency parser\footnote{https://spacy.io/api/dependencyparser}. Then, we classified a sentence as passive if it contained either an 'agent' relation or an 'nsubjpass' relation.

\textbf{Mean Sentence Count with Past Tense}: We counted the number of sentences in a given paragraph that were written in the past tense. After extracting the sentences for a given paragraph, we used POS tags to determine whether a sentence was in the past tense. Specifically, we checked if the sentence's POS tags included "VBD" or "VBN" and classified the sentence as past tense accordingly.

\subsection{$F^2$: \textbf{Punctuation usage}}

Under punctuation usage, we calculated the average occurrence of "$^^21$" "\#", "\textquotesingle", and the Oxford comma per paragraph. We divided the input news article into paragraphs, following the same approach as the previous feature extractions. For each paragraph, we determined the frequency of the mentioned punctuation.

\subsection{$F^3$: \textbf{Formatting standard violations}}

Here, we identify and tally violations related to date, time, and number formats in accordance with the AP standards.

\textbf{Date format violations}: To detect date format violations, we utilize the datefinder library\footnote{https://pypi.org/project/datefinder/} to extract date elements from the input news article. We then verify if the date adheres to the standard format specified by AP standards. Specifically, we ensure that the day is written in full without abbreviations (e.g., "Monday," "Tuesday," etc.), and that the month is represented correctly. For instance, when a month is used with a specific date, we expect abbreviations like "Jan.," "Feb.," "Aug.," "Sept.," "Oct.," "Nov.," and "Dec." When a phrase only lists the month and year, the month should be spelled out, and there should be no comma separating the month and year.

\textbf{Time format violations}: In assessing time format violations, we verify if time phrases in the news article adhere to the AP standards. This entails using lowercase "a.m." and "p.m." with periods and ensuring that numerals precede the time. If a time phrase does not conform to these standards, it is considered a format violation. Similar to the date format analysis, we employ the datefinder library to extract time phrases from the text.

\textbf{Number format violations}: We identify number format violations when numbers in an article fail to comply with the AP standards. According to these standards, numbers from zero to nine should be spelled out, while numerals should be used for 10 and above.

Altogether we collected \textbf{14} journalism features through the above extractors. 

\section{Baselines}
\label{sec:appendix_base}

\textbf{GLTR} ~\cite{gehrmann2019gltr}: This approach utilizes a proxy language model (PLM) to calculate the log probabilities of tokens in the input text. The authors then incorporate a set of statistical scores to predict the label, including average log probability, average token rank, token log-rank, and predictive entropy. For example, a higher average log probability of input indicates AI generation. The second and third scores share a similar assumption, where lower average ranks in input suggest AI-generated text. The final score is based on the hypothesis that AI-generated text tends to have lower entropy. Our paper reports the average performance across all the trials based on the scores mentioned above.

\noindent \textbf{DetectGPT}~\cite{mitchell2023detectgpt}: This approach also utilizes a proxy PLM to calculate log probabilities for individual tokens. However, its decision process involves comparing the log probability of the original input text with the log probability of a set of perturbed versions of the input text. These perturbations are generated using the T5-base. The authors hypothesize that if the difference in log probabilities between the original text and the perturbed text is consistently positive, then it is likely that an AI model generated the input text.

\noindent \textbf{OpenAI-GPT2 detector}: This detector is a RoBERTa model fine-tuned on the GPT-2-output dataset\footnote{https://github.com/openai/gpt-2-output-dataset} which consists of 250K documents from the WebText dataset ~\cite{radford2019language} and 500K GPT2 generated data. We incorporate two variants of this model: 1) OpenAI$_{Zero}$ - off-the-shelf model without any additional finetuning on AI-generated news detection task and 2) OpenAI$_{FT}$ - off-the-shelf model further finetuned on training datasets used for AI generated news detection task. 

\section{Implementation Details}
\label{sec:appendix_imp}

Apart from the implementation details discussed in Section \ref{det_setup}, another crucial aspect to consider is the hyperparameters of the Guidance Head and Classification Head, including layer sizes.

\textbf{Guidance Head}: As described in the methodology section, the Guidance Head comprises one hidden layer that maps the input $[B_{d}^{CLS}, J_{n}]$ to a higher-dimensional feature space. Consequently, we opted for a larger hidden layer size compared to the input size $d+n$ (Base detector hidden size + journalism feature vector size). In the case of RoBERTa-base $d=768$, where $d+n=768+14=782$, we found that a Guidance Head layer size of 1024 yielded the best performance. Additionally, we used an output layer of size 256 for optimal results. To summarize, the layers of the Guidance Head are structured as 782 $\rightarrow$ 1024 $\rightarrow$ 256.

\textbf{Classification Head}: The decision regarding the layer size for the Classification Head was straightforward. Essentially, starting from the output size of the Guidance Head, our objective was to obtain the final prediction for the two classes. The layer sizes that achieved the best performance were 256 $\rightarrow$ 32 $\rightarrow$ 2.

\section{Additional Experiment Results}
\label{sec:appendix_results}

\subsection{Journalism Feature Importance}

We conducted a study to better understand the significance of journalism features in AI news detection with the help of a Shapley Additive Explanations (SHAP) \cite{lundberg2017unified} explainer on the logistic regression classifier. In section \ref{re_dis}, we only presented the SHAP plots for the GPT3 detection task for brevity reasons. Therefore, here we present the additional SHAP plots related to other PLMs detection tasks. 


\begin{figure}[h]
     \centering
     \includegraphics[width=0.45\textwidth]{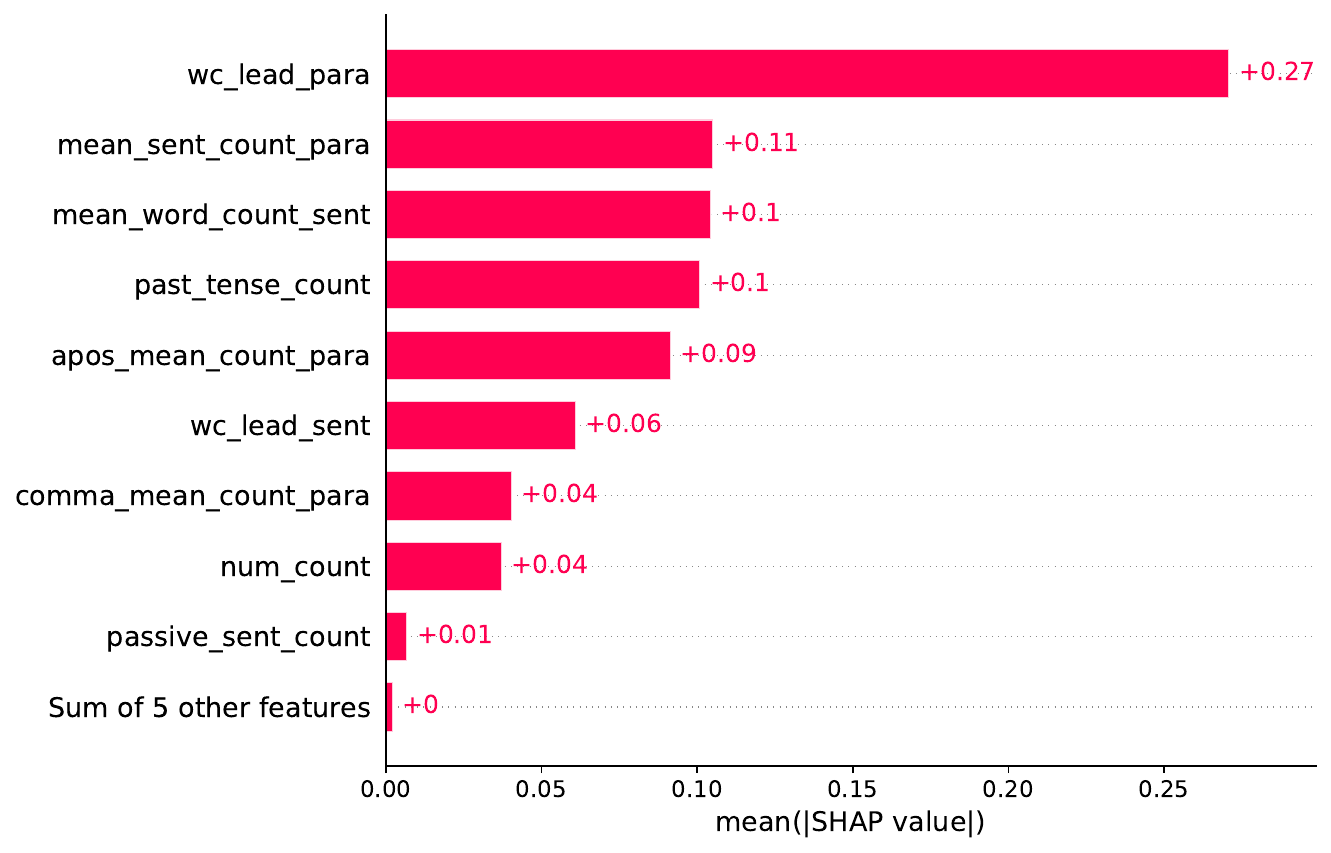}
     \caption{SHAP values to estimate journalism features importance - CTRL}
\label{fig:feat_impo_ctrl}
\end{figure}

\begin{figure}[h]
     \centering
     \includegraphics[width=0.45\textwidth]{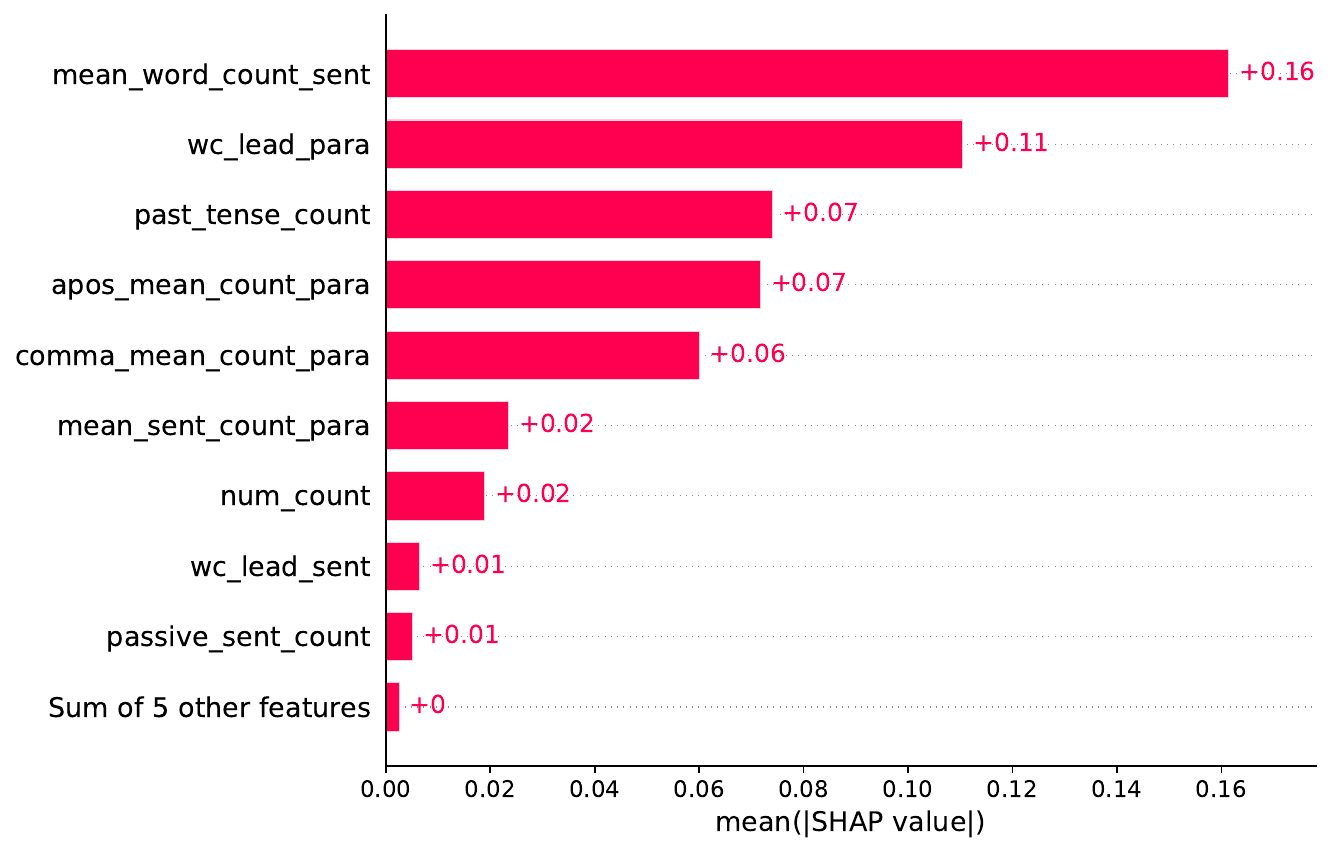}
     \caption{SHAP values to estimate journalism features importance - GROVER}
\label{fig:feat_impo_grover}
\end{figure}

\begin{figure}[h]
     \centering
     \includegraphics[width=0.45\textwidth]{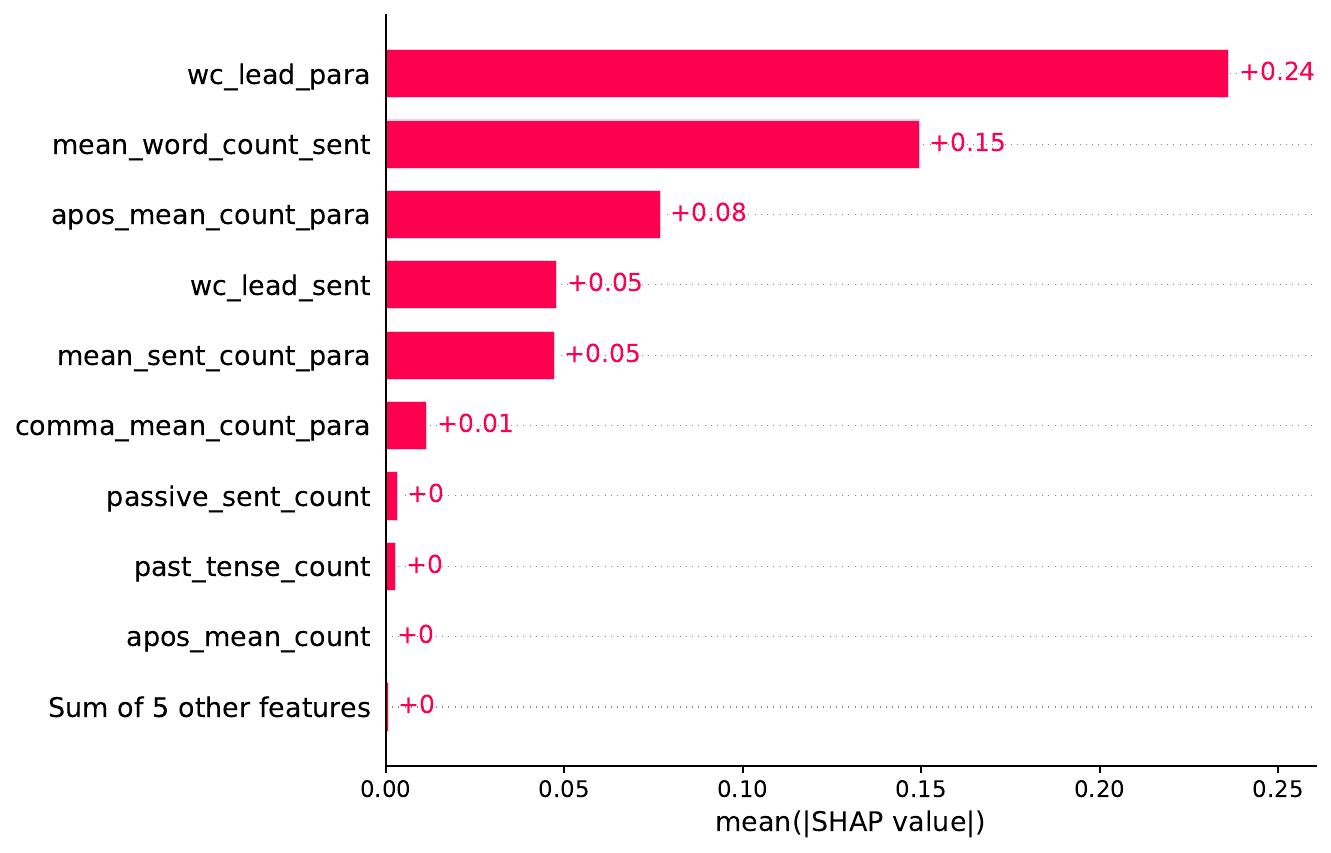}
     \caption{SHAP values to estimate journalism features importance - GPT2}
\label{fig:feat_impo_gpt2}
\end{figure}

\begin{figure}[h]
     \centering
     \includegraphics[width=0.45\textwidth]{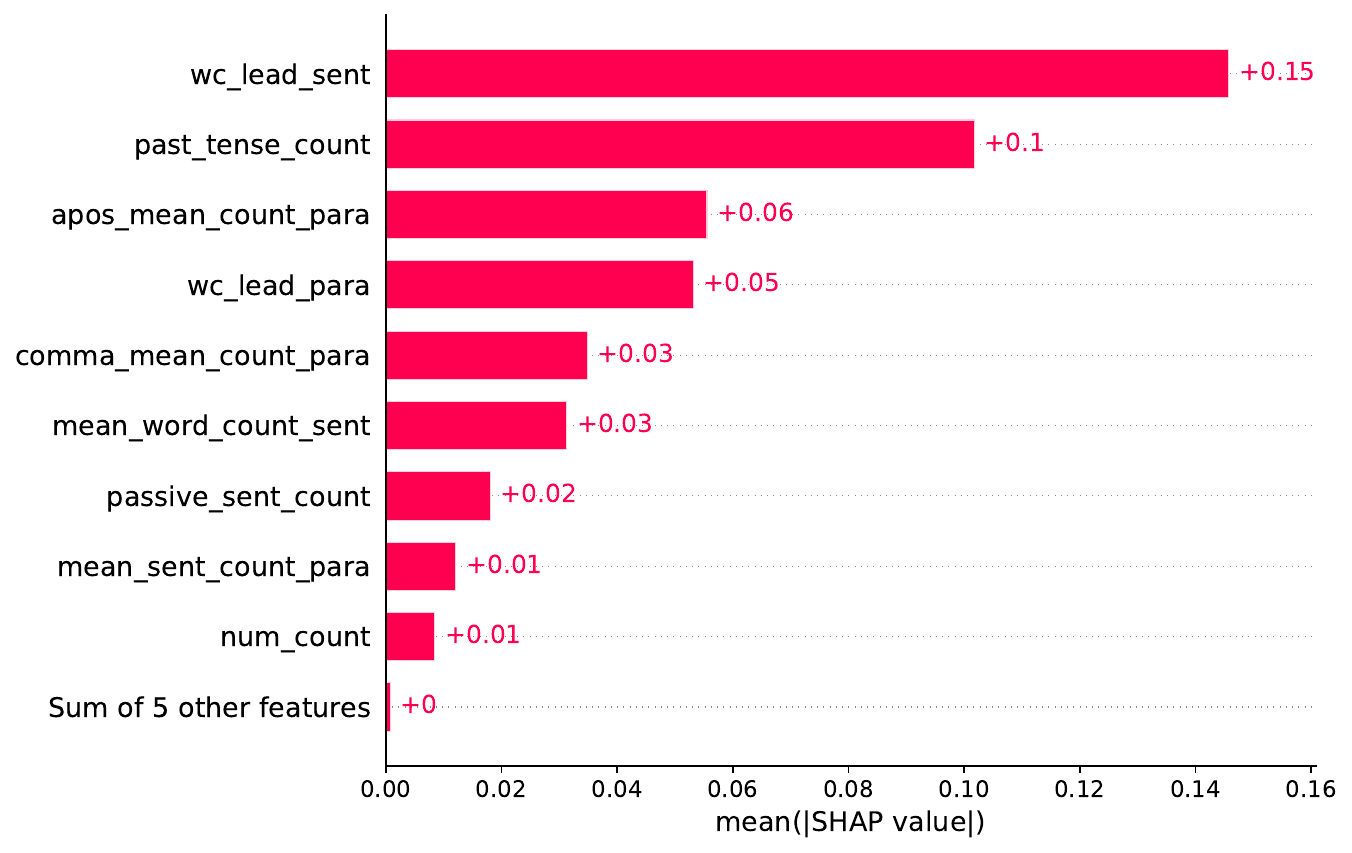}
     \caption{SHAP values to estimate journalism features importance - PPLM$_gpt2$}
\label{fig:feat_impo_pplm}
\end{figure}

\begin{figure}[h]
     \centering
     \includegraphics[width=0.45\textwidth]{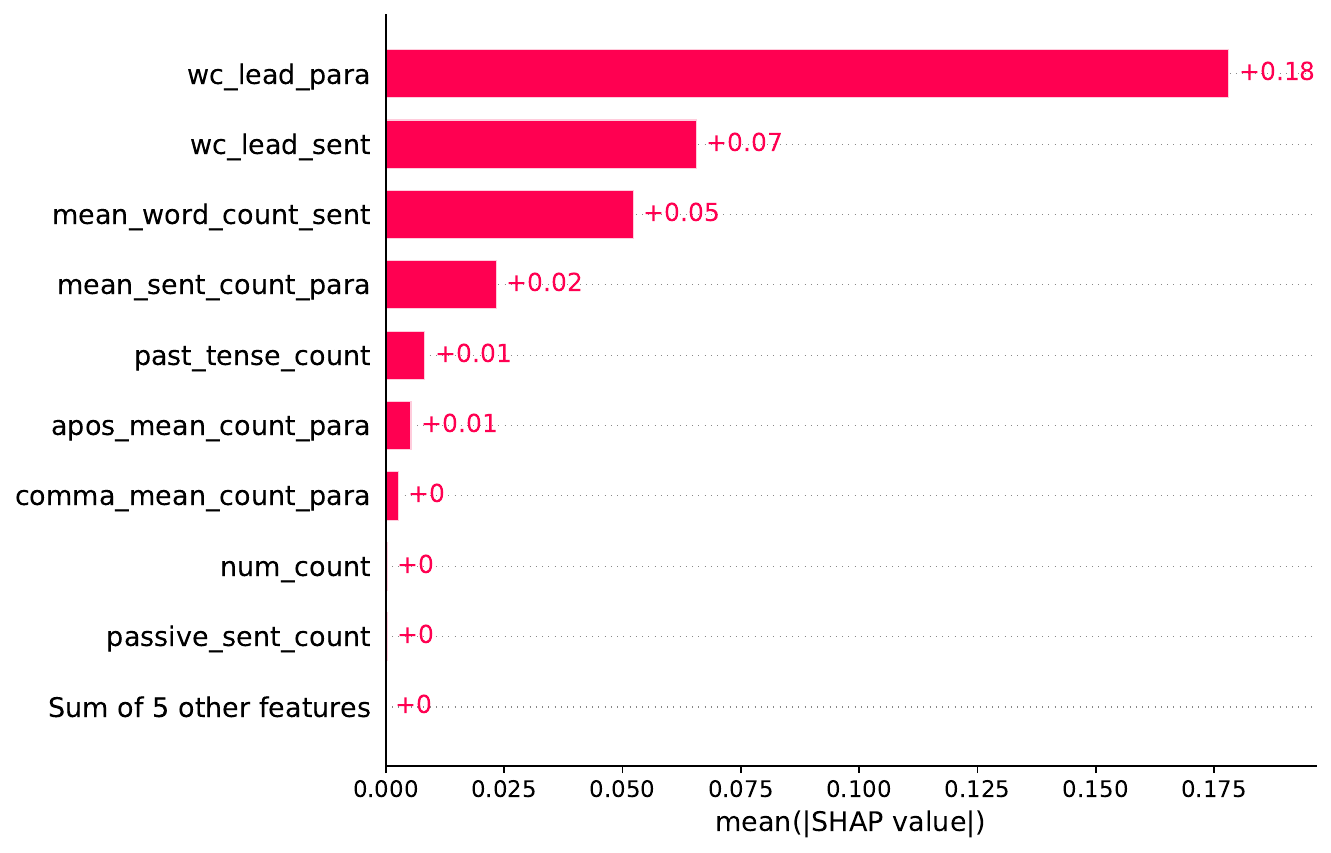}
     \caption{SHAP values to estimate journalism features importance - ChatGPT}
\label{fig:feat_impo_chatgpt}
\end{figure}

We see that almost all the SHAP plots agree on the importance of certain features, such as leading paragraph and or sentence word count, apostrophe usage, and past tense usage.

\subsection{Adversarial Robustness Results}

In section \ref{rq2}, we discussed our experiments on evaluating the adversarial robustness of AI news detectors. We conducted two types of attacks on the detectors: character-level attacks involving Cyrillic injection and word-level attacks involving paraphrasing. We only report the detection performance on GPT3 and ChatGPT data for brevity in section  \ref{rq2}. Therefore, here we present the results of the rest of the PLMs of our study. Table \ref{tab:robust_results_apendix1} and Table \ref{tab:robust_results_apendix2} show the detectors' performance (AUROC) difference before and after each attack. 

\begin{table}[]
\caption{Detector performance change after the attack (AUROC before the attack - AUROC after the attack). Bold shows the lowest AUROC difference within each column (detector-attack combination)}
\label{tab:robust_results_apendix1}
\centering
\begin{tabular}{|l|ll|ll|}
\hline
Generator   $\rightarrow$          & \multicolumn{2}{l|}{Grover}                                          & \multicolumn{2}{l|}{CTRL}                                            \\ \hline
Attack    $\rightarrow$     & \multicolumn{1}{l|}{\multirow{2}{*}{Para.}} & \multirow{2}{*}{Cyri.} & \multicolumn{1}{l|}{\multirow{2}{*}{Para.}} & \multirow{2}{*}{Cyri.} \\ \cline{1-1}
Detector  $\downarrow$     & \multicolumn{1}{l|}{}                       &                        & \multicolumn{1}{l|}{}                       &                        \\ \hline
GLTR          & \multicolumn{1}{l|}{\textbf{0.035}}         & \textbf{0.038}         & \multicolumn{1}{l|}{0.233}                  & 0.186                  \\ \hline
DetectGPT     & \multicolumn{1}{l|}{0.127}                  & 0.086                  & \multicolumn{1}{l|}{0.248}                  & 0.195                  \\ \hline
OpenAI$_Zero$ & \multicolumn{1}{l|}{0.233}                  & 0.169                  & \multicolumn{1}{l|}{0.229}                  & 0.175                  \\ \hline
OpenAI$_FT$   & \multicolumn{1}{l|}{0.144}                  & 0.113                  & \multicolumn{1}{l|}{0.162}                  & 0.106                  \\ \hline
{\m}             & \multicolumn{1}{l|}{0.082}                  & 0.054                  & \multicolumn{1}{l|}{\textbf{0.074}}         & \textbf{0.031}         \\ \hline
\end{tabular}
\end{table}

\begin{table}[]
\caption{Detector performance change after the attack (AUROC before the attack - AUROC after the attack). Bold shows the lowest AUROC difference within each column (detector-attack combination)}
\label{tab:robust_results_apendix2}
\centering
\begin{tabular}{|l|ll|ll|}
\hline
Generator         $\rightarrow$           & \multicolumn{2}{l|}{PPLM$_{gpt2}$}                                            & \multicolumn{2}{l|}{GPT2}                                            \\ \hline
Attack      $\rightarrow$           & \multicolumn{1}{l|}{\multirow{2}{*}{Para.}} & \multirow{2}{*}{Cyri.} & \multicolumn{1}{l|}{\multirow{2}{*}{Para.}} & \multirow{2}{*}{Cyri.} \\ \cline{1-1}
Detector    $\downarrow$           & \multicolumn{1}{l|}{}                       &                        & \multicolumn{1}{l|}{}                       &                        \\ \hline
GLTR                  & \multicolumn{1}{l|}{0.124}                  & 0.082                  & \multicolumn{1}{l|}{0.090}                  & 0.054                  \\ \hline
DetectGPT             & \multicolumn{1}{l|}{0.083}                  & 0.050                  & \multicolumn{1}{l|}{0.083}                  & 0.029                  \\ \hline
OpenAI$_{Zero}$         & \multicolumn{1}{l|}{0.187}                  & 0.123                  & \multicolumn{1}{l|}{0.237}                  & 0.170                  \\ \hline
OpenAI$_{FT}$           & \multicolumn{1}{l|}{0.140}                  & 0.092                  & \multicolumn{1}{l|}{0.181}                  & 0.137                  \\ \hline
{\m} & \multicolumn{1}{l|}{\textbf{0.082}}         & \textbf{0.042}         & \multicolumn{1}{l|}{\textbf{0.073}}         & \textbf{0.040}         \\ \hline
\end{tabular}
\end{table}

Table \ref{tab:robust_results_apendix1} and Table \ref{tab:robust_results_apendix2} hold similar observations as we presented with GPT3 and ChatGPT3 data in section \ref{rq2}. Almost every SOTA baseline detector we have considered is susceptible to adversarial attacks. On average, the performance of the detectors dropped by at least 10-20\%. However, in some cases, we observed a low attack success rate with the GLTR and DetectGPT. However, the observation of low attack success is meaningless as these models had a near-random guess ($\approx0.5$ AUROC) performance before the attack. 

Similar to GPT3 and ChatGPT detection, we can observe that {\m} is quite resilient to adversarial attacks across other PLM generators, with an average performance drop of only 7\%. It is again evident that this robustness is due to the journalism features employed by \m. For example, OpenAI{\tiny FT}, which shares the same PLM architecture and training data for detection as {\m}, has an average performance drop of nearly 15\%. 

\label{sec:appendix_exp}

\end{document}